\documentclass[letterpaper, 10 pt, conference]{ieeeconf}  % ICRA format

\IEEEoverridecommandlockouts
\usepackage{cite}
\usepackage{amsmath,amssymb,amsfonts}
\usepackage{algorithmic}
\usepackage{graphicx}
\usepackage{textcomp}
\usepackage{xcolor}
\usepackage{hyperref}
\usepackage{flushend} % Balance columns on last page
\def\BibTeX{{\rm B\kern-.05em{\sc i\kern-.025em b}\kern-.08em
    T\kern-.1667em\lower.7ex\hbox{E}\kern-.125emX}}

\usepackage{todonotes}

\usepackage{array}
\usepackage{lipsum}
\usepackage{subcaption} % ythis makes table captions wrong
\usepackage{float}
\usepackage{svg}

\usepackage{booktabs}
\usepackage{balance}

\usepackage{hyperref}
\usepackage{siunitx} % SI units
\usepackage[font=small]{caption}

\begin{document}

\title{\LARGE \bf LiSTA: Geometric Object-Based Change Detection\\ in Cluttered Environments\\
\thanks{This work has been carried out within the framework of the EUROfusion Consortium, funded by the European Union via the Euratom Research and Training Programme (Grant Agreement No 101052200 — EUROfusion). Views and opinions expressed are those of the author(s) do not necessarily reflect those of the European Union or the European Commission. Neither the European Union nor the European Commission can be held responsible for them. Additional funding from an EPSRC IAA grant (EP/X525777/1) and a Royal Society University Research Fellowship (Fallon).
For the purpose of open access, the authors have applied a Creative Commons Attribution (CC BY) license to any Accepted Manuscript version arising.}
}

\author{Joseph Rowell$^1$, Lintong Zhang$^1$, and Maurice Fallon$^1$%
\thanks{\hspace{-1em}$^{1}$Oxford Robotics Inst., Dept. of %
Engineering Science, Uni. of Oxford, UK. \newline%
\texttt{\{joseph, lintong, mfallon\}@robots.ox.ac.uk}\newline%
}}

\maketitle

\begin{abstract}
We present LiSTA (LiDAR Spatio-Temporal Analysis), a system to detect probabilistic object-level change over time using multi-mission SLAM.  
Many applications require such a system, including construction, robotic navigation, long-term autonomy, and environmental monitoring. We focus on the semi-static scenario where objects are added, subtracted, or changed in position over weeks or months. Our system combines multi-mission LiDAR SLAM, volumetric differencing, object instance description, and correspondence grouping using learned descriptors to keep track of an open set of objects. Object correspondences between missions are determined by clustering the object's learned descriptors. We demonstrate our approach using datasets collected in a simulated environment and a real-world dataset captured using a LiDAR system mounted on a quadruped robot monitoring an industrial facility containing static, semi-static, and dynamic objects. Our method demonstrates superior performance in detecting changes in semi-static environments compared to existing methods.

\end{abstract}
% \begin{IEEEkeywords}
% Object detection, segmentation, categorization;
% Field Robots;
% Mapping;
% Robotics and Automation in Construction
% \end{IEEEkeywords}

\section{Introduction}

Change detection is an important capability for autonomous robots doing environmental monitoring, infrastructure management, and disaster response. LiDAR and cameras are used when monitoring and assessing structural changes and damages to buildings, roads, bridges, and infrastructure, ensuring maintenance and safety \cite{Hao2020RemoteSensing, Lorenz2017ChangeDetection, MarinelliPB18}.
In addition, the use of 3D imaging has the potential to enhance the effectiveness and surpass some of the restrictions of conventional 2D image-based change detection \cite{Qin20163DApplications}. Camera images capture fine visual detail which enable corrosion detection and instrument reading \cite{Jian2020InstrumentReading}, while LiDAR is complementary by accurately detecting any physical changes.

The DARPA Subterranean Challenge demonstrated that 3D mapping systems are now highly accurate and increasingly mature \cite{Chung2023IntoChallenge}.
Acquiring large, long-term mapping datasets is becoming increasingly feasible, however to automate change detection one must consider the noisy and incomplete observations from mobile robot surveys of a facility. Beyond the process of detecting change, it is advantageous to identify which objects have moved, and to where, providing valuable information for the surveyor.
%The use of LiDAR technology to carry out \emph{3D physical change detection} is motivated because purely photogrammetry-based approaches cannot perceive subtle geometric changes.

\begin{figure}
    \centering
    \includegraphics[width=\columnwidth]{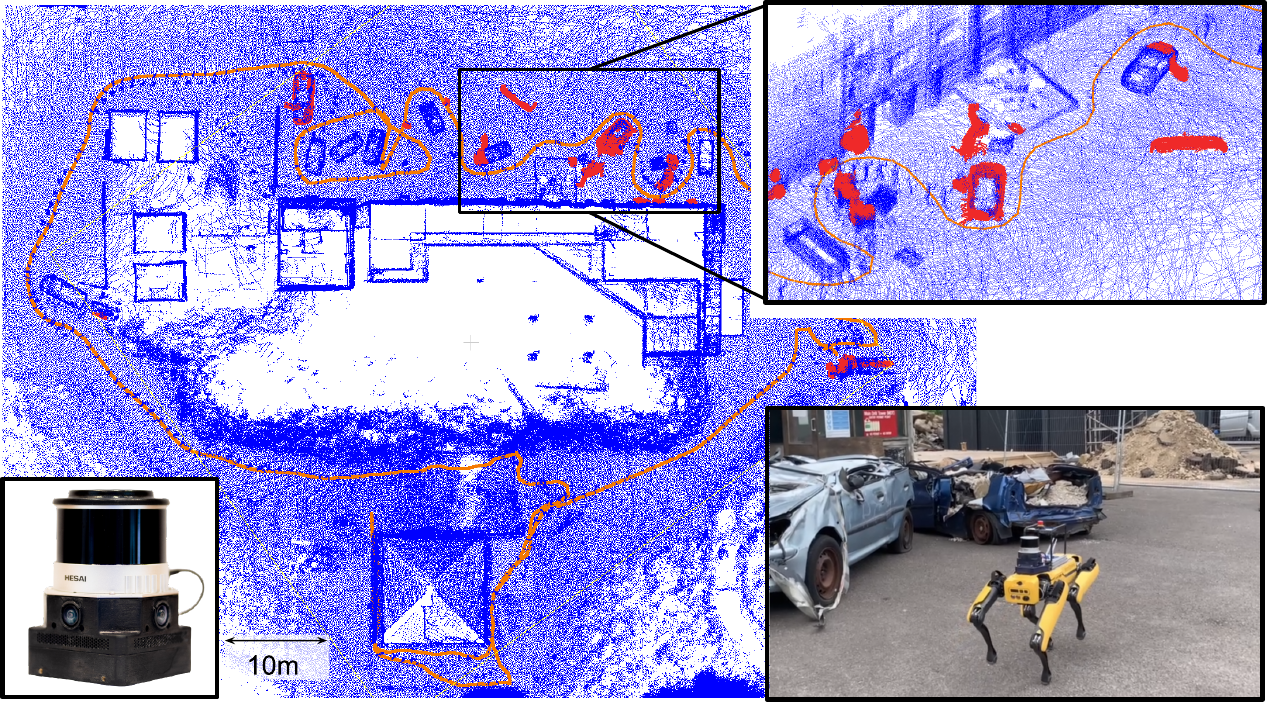}
    \caption{Top down view of 3D maps acquired by an autonomous Spot inspecting Fire Service College, static map in blue; with changes identified between missions A and B in red. Photos show our in-house sensor suite \emph{Frontier} and Spot quadruped.}
    \label{fig:fsc_teaser}
\end{figure}

In this work, we present a novel LiDAR-based change detection framework that estimates object correspondences using a learning-based object descriptor. Our method uses octrees to represent the environment volumetrically. By differencing octrees from different robot missions, we can identify clusters corresponding to changes in the environment.
%We present a qualitative and quantitative evaluation
%To detect which objects and moved and to where, we developed unsupervised object discovery and matching method \mfallon{im confused about this:} that determines correspondences with discovered change and a label-less open set of objects. We used a convolutional neural network (CNN) to generate descriptive embeddings for these changed objects and these embeddings allowed us to establish correspondences between objects observed across different missions, providing a robust method to track objects over time.
To demonstrate the potential for real-world applications, we collected datasets with an autonomous Boston Dynamics Spot robot patrolling an industrial environment, see Fig. \ref{fig:fsc_teaser}. We also quantitatively evaluated the approach on simulated datasets.

Our approach works without the need for pre-training on a set of specific objects. This is particularly important when operating in cluttered industrial environments containing unusual plant equipment, atypical to pre-trained model classes. The key contributions are:
\begin{itemize}
    \item A holistic approach to LiDAR object-level change detection, combining multi-mission SLAM, advanced data representation, segmentation, and deep learning techniques.
    \item A change detection algorithm that allows us to segment and classify discrete objects that are repositioned between multiple missions, without requiring pre-training on a closed set of objects.
    \item A correspondence grouping method and a confidence metric that offers
    a solution to quantify uncertainty when classifying the changed objects.
    \item The multi-mission simulated LiDAR change detection dataset\footnote{\url{https://ori.ox.ac.uk/labs/drs/datasets-drs/}} used in our evaluations, which could be useful to others researching object-level change detection.
\end{itemize}

\section{Related Work}

\subsection{Change Detection}
    Generally, the two approaches taken to detect changes at the object-level
    are direct model comparison and classification-based comparison
    \cite{Singh1989DigitalData}. Direct comparison detects changes directly
    from the raw multi-temporal data, while classification-based comparison
    classifies objects and subsequently tracks their movements.
    Classification-based methods that use learning, such as HGI-CD
    \cite{Ku2021SHRECScenes} can achieve remarkable performance, although one
    significant limitation is the challenge of class imbalance; this can lead
    to a bias towards a majority class and poor performance in minority
    classes. Our method can overcome the class imbalance problem by introducing
    a confidence metric for label-less classification, as well as the ability
    to generalize without pre-training on specific object categories.
    % such as \cite{Gallagher2009GATMO:Objects, Biswas2002TowardsRobots,
    %FehrTSDF-basedDiscovery} ...

     Methods carrying out direct comparison, such as Saarinen et al.
     \cite{Saarinen2012IndependentEnvironment}, use an occupancy grid with the
     assumption that each voxel is an independent Markov chain, while
     Andreasson \textit{et al.} \cite{Andreasson2007HasRobots} used a 3D
     version of the Normal Distribution Transform (NDT) to detect not only
     geometric but also RGB variations between a reference mission and a new
     observation. Fehr \textit{et al.} \cite{FehrTSDF-basedDiscovery} used
     RGB-D cameras for TSDF generation, and performed unsupervised object
     discovery. Although RGB-D cameras are low cost, they have limited depth
     perception at long ranges and are sensitive to lighting conditions
     \cite{Debeunne2020AMapping}, making them less useful for our application;
     these issues are not present in LiDAR.
    LiSTA acts by performing unsupervised classification of changed objects,
    determined directly from raw data. We offer a robust solution for accurate
    change detection, even when the changed objects are atypical, such as those
    in a cluttered industrial environments.

\subsection{Correspondence Grouping}
    Traditionally, object classification or correspondence grouping methods use
    a pre-trained neural network, trained on a library of known and labeled
    objects
    \cite{Alhamzi20153DLibrary,Jiang2018PointSIFT:Segmentation,Zhang2020DeepSegmentation},
     however, these often don't generalize to other domains. Since point cloud
    annotation is time-consuming and laborious, these pre-trained model
    datasets are often severely constrained in data size and data diversity. In
    realistic situation we often find objects which are atypical to training
    datasets, causing erroneous classification and labeling, for example,
    labeling data to classify all the objects found in a nuclear facility would
    be labour intensive and impractical.  We perform 3D object recognition
    through correspondence grouping, in order to cluster the set of
    point-to-point correspondences obtained after a 3D descriptor matching
    stage into model instances that are present in the current scene.  Both
    data-driven and metric-driven approaches are explored. Metric-driven
    methods, such as RIFT (Rotation-Invariant Feature Transform)
    \cite{Arbeiter2012EvaluationClouds} or SHOT (Signature of Histograms of
    OrienTations) \cite{Aldoma2012Tutorial:Estimation}, calculate 3D feature
    descriptors for downsampled point cloud objects and match them against a
    database of object descriptors, generating a list of instance to instance
    correspondences without the requirement of a neural network with labeled
    training data.

    RIFT is a descriptor designed to be invariant to the rotation of the
    object, and operates by partitioning a local region around each point of
    interest into multiple concentric rings and quantizing the point
    distribution within these rings. The resulting histogram encodes
    information about the spatial distribution of points, which makes it robust
    to changes in viewpoint. Whereas SHOT encodes the geometric properties of a
    point by computing a histogram of surface normal orientations within its
    local neighborhood.
    However, these traditional surface geometry point cloud descriptors have
    limitations in terms of computational efficiency and the ability to capture
    complex patterns. Furthermore, when matching objects in a large database,
    efficiently comparing the RIFT or SHOT descriptors can become a bottleneck,
    especially for dense point clouds.

    In contrast, learning-based descriptors, such as
    PointNet~\cite{QiPointnet:Segmentation}, can adapt to the complexity of the
    data. They learn discriminative features directly from the point cloud
    data, making them effective at capturing intricate patterns, which can be
    challenging for traditional descriptors. However, learning-based methods
    struggle to run in real time for mobile robotic applications.
    %\mfallon{this reads like `we do grouping to achieve grouping'. rephrase}
    Our previous work InstaLoc ~\cite{ZhangInstaLoc:Learning}, can localize an
    individual LiDAR scan within a prior map by matching object instances in
    the query scan with those in the map. It uses a fast and efficient
    descriptor network to describe each object.
    In this work, we adapt the descriptor network to learn the objects' 3D
    features and group similar objects into a cluster. Details are discussed in
    Sec \ref{sec:inst_descriptor}.
    %We aimed to perform 3D object recognition in order to cluster the set of
    %point-to-point correspondences obtained after the 3D descriptor matching
    %stage into model instances that are present in the current scene.

    \begin{table}
        \caption{\centering Octree resolutions  (Res.) in state-of-the-art
        implementations.} \label{tab:octomap_resolution}
        % \centering
        \resizebox{\columnwidth}{!}{\begin{tabular}{c  c  c  c  c}
             \toprule
             \textbf{Work} & \textbf{Res. (cm)} & \textbf{Environment} &
             \textbf{LiDAR} & \textbf{Range (m)}\\ [0.5ex]
             \midrule
             \cite{ArnesenOctoMap:Systems} & 5 & Indoor &  SICK LMS & 25\\
             %wurm \textit{et al.}
             \cite{Wang2020ActivelyRobot} & 5 & Outdoor &  Velodyne VLP-16  &
             100\\  %wang \textit{et al.}
              \cite{Reijgwart2020Voxgraph:Submaps} & 5 & Outdoor & Ouster OS1
              &  120\\ % Reijgwart \textit{et al.}
             \cite{ReijgwartEfficientSystems} & 2, 5 &In \& Outdoor & Velodyne
             VLP-16 & 100\\
             Ours & 5 & In \& Outdoor & Hesai XT32  & 120 \\
              % reigjwart \textit{et al.}
             \bottomrule
        \end{tabular}}
    \end{table}

    %% fullsize original table
    % \begin{table*}[htbp]
    %     \caption{octree resolutions in state-of-the-art, RES: resolution
    %\lintong{This table may not be necessary.}} \label{tab:octomap_resolution}
    %     \centering
    %     \begin{tabular}{c  c  c  c  c }
    %          \hline
    %          \textbf{Paper} & \textbf{RES (cm)} & \textbf{Environment} &
    %\textbf{LiDAR Sensor} & \textbf{System} \\ [0.5ex]
    %          \hline
    %          Wurm et al \cite{ArnesenOctoMap:Systems} & 5 & Indoor &  SICK
    %LMS &  Intel E8600 CPU\\ %wurm et al
    %          \hline
    %          Wang et al \cite{Wang2020ActivelyRobot} & 5 & Outdoor &
    %Velodyne VLP-16  & CPU \\  %wang et al
    %          \hline
    %           Reijgwart et al \cite{Reijgwart2020Voxgraph:Submaps} & 5 &
    %Outdoor & Ouster OS1   & Intel i7-8650 CPU \\ % Reijgwart et al
    %            & & & \& RealSense D415 RGB-D & \\
    %          \hline
    %          Reijgwart et al \cite{ReijgwartEfficientSystems} & 2, 5 &Indoor
    %\& Outdoor & Velodyne VLP-16 &  Intel i9-9900K CPU \\
    %          \hline
    %          Ours & 5 & Indoor \& Outdoor & Hesai XT32  & Intel i7-10875H CPU
    %\\
    %          [1ex] % reigjwart et al
    %          \hline

    %     \end{tabular}
    % \end{table*}

\section{Method}
\begin{figure}
        \centering
        \includegraphics[width=\columnwidth]{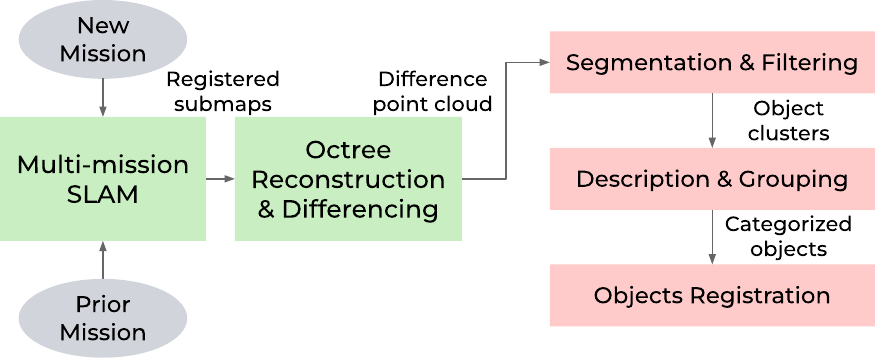}
        \caption{LiSTA change detection method overview. Five modules process the sensor data to output the classified changed objects intermission.}
        \label{fig:alg_diagram}
    \end{figure}

    The system overview is presented in Fig. \ref{fig:alg_diagram}. It first uses multi-mission pose-graph LiDAR simultaneous localization and mapping (SLAM) to create co-registered point cloud maps. Octrees are then produced and differenced to determine occupancy probabilities. Voxels containing change are projected back into raw point clouds. Discrete objects are then segmented, and finally correspondences are established between the discovered objects.

    % \begin{figure}
    %     \centering
    %     \includegraphics[width=0.6\columnwidth]{figures/sensor_suite.pdf}
    %     \caption{In-house sensor suite, \emph{Frontier}, used to collect the real-world dataset. It consists of a Hesai XT32 LiDAR, three Sevensense Alphasense cameras and an IMU.}
    %     \label{fig:sensor_suite}
    % \end{figure}

    %mfallon{This is confusing: it say 4 modules but shows 5 modules - and in particular `Multi-mission Registration' doesnt say 'VILENS SLAM'}
    %\mfallon{ I think just change it to `Multi-Mission SLAM' is sufficient. And replace VILENS SLAM with the same phrase. Also the shared colours implie a connection to Fig 3 which could be confusing.}

    %\mfallon{the above summary is also presented in the introduction. Can you please remove nearly the entirety of the paragraph before the list of contributions?} -- done

    \subsection{3D Reconstruction}
    \subsubsection{Multi-Mission Point Cloud SLAM} \label{sec:multi-mission}
        Our approach begins with the acquisition of multiple SLAM maps using our VILENS SLAM system~\cite{Camurri2022HILTISLAM} (which won the 2021 Hilti SLAM Challenge), this system uses LiDAR inertial odometry as a basis and a pose-graph optimization based on iSAM2~\cite{Dellaert2017} with intra-mission loop closures detected during operation. Point clouds obtained in the different missions need to be aligned in a common reference frame to facilitate direct comparison. To ensure map consistency we do not directly register the global point cloud maps output by each SLAM mission as that would result in erroneous change being detected (see an example in Sec. \ref{sec:fsc_change_detection}). Instead we first jointly align the multi-mission pose-graphs.

    \begin{figure}
        \centering
        \includegraphics[width=0.9\columnwidth]{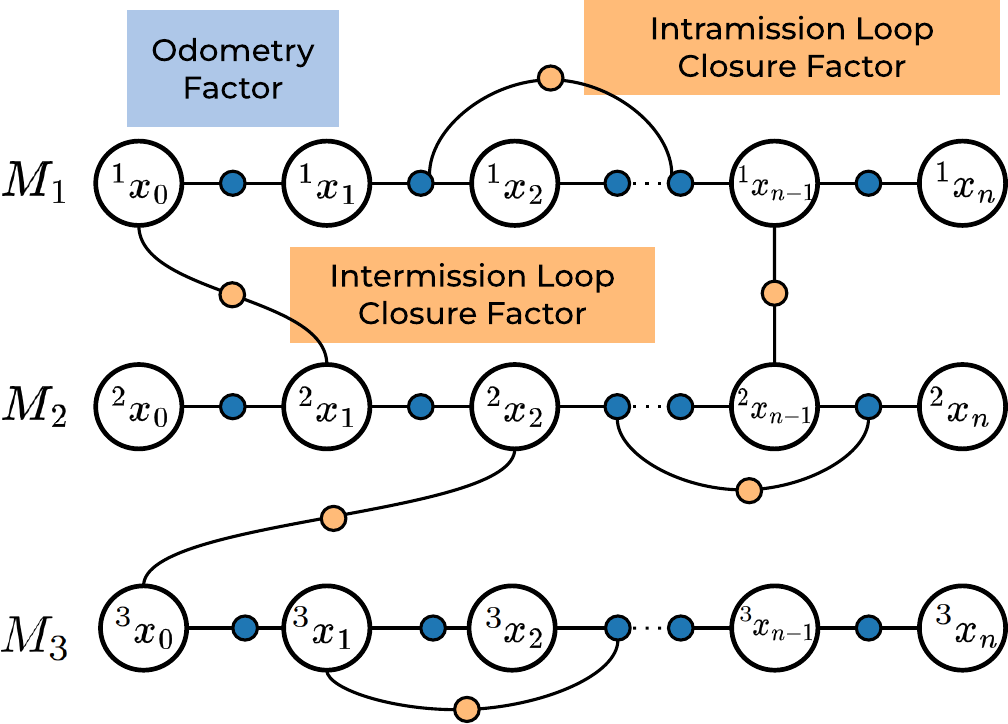}
        \caption{Multi-Mission SLAM registration factor graph structure: It identifies loop closures between missions ($M_1$, $M_2$ \& $M_3$) and jointly optimizes their pose-graphs.}
        \label{fig:factor_graph}
    \end{figure}

        To do this we employ multi-mission pose-graph SLAM as illustrated in Fig~\ref{fig:factor_graph}. First, candidates for intermission loop closure are proposed by matching ScanContext \cite{scan_context} descriptors between the individual LiDAR scans from each mission with each loop closure candidate then refining using iterative closest point (ICP) registration to form the factor. The missions can then be jointly optimized when loop closure constraints are identified between them. Finally, the combined pose-graph is optimized to obtain the final trajectory for the missions in a common coordinate frame. This assumes that the individual missions overlap with at least one other mission.
        %\mfallon{This is all about multi-session SLAM but we only have two sessions to talk about - which isn't really multi. Just Sayin'} -

        This approach allows the creation of a unified ``4-D'' map that represents the entire environment explored by the robot across the different missions, and enables direct change detection by comparing the point cloud map between different missions in the same reference frame. This method of registration is also robust to geometric change intermission as loop closures are inherently local.
        % \mfallon{the concept of 4D mapping is not demonstrated here - we only have a pair missions. In the video it would be good to show the 5 Mickie lane maps stacked on top of one another - except offset in the Z-axis}
        % OK. There are actually multiple missions (5) for fsc exp but the paper just shows 2 missions LZ

    \subsubsection{Octree Reconstruction and Differencing}
        The registered local maps were then used to generate an octree, allowing efficient and accurate volumetric differencing. The octree data structure, from OctoMap \cite{Hornung2013OctoMap:octrees}, was chosen for change detection because it can finely represent occupied and unoccupied 3D space through ray casting from the sensor to observed points, distinguishing between unobserved and unoccupied areas.

        % \mfallon{this detail is unnecessary - please make this half as long. it is 10yrs old: is a probabilistic 3D representation chosen for our change detection due to its ability to represent both occupied and unoccupied space at a fine-grained voxel level, via casting rays from the sensor to the observed points and updating occupancy accordingly. Ray casting allows us to differentiate between unobserved and unoccupied space.}

        The resolution of the octree was chosen to be $5cm$ to match the cloud density produced by our Hesai XT32 LiDAR and is consistent with other works, Tab. \ref{tab:octomap_resolution}.
        % octrees employ a hierarchical tree-like data structure to partition 3D space into octants, allowing for adaptive voxel resolution.

        % \mfallon{you need need to say the octree threshold value}
        To identify the differences between the octrees, we iterated through the overlapping octree nodes and compared the occupancy probabilities of the corresponding voxels.
        To distinguish changed from static elements, we set an occupancy threshold of $0.5$. Voxels with occupancy probabilities exceeding this threshold were considered changed, indicating areas where changes had occurred between missions. Conversely, voxels with occupancy probability difference below the threshold were classified as static, representing parts of the environment that remained unchanged.

        %The resolution of the octree was chosen to be $5cm$, as it matches the resolution of the Hesai Pandar LiDAR in use.

        %The resolution of an octree refers to the size or granularity of the cubic regions that it uses to partition space. Higher resolution means smaller cubic regions, while lower resolution means larger cubic regions.
        %The resolution of the octree was chosen to be 5cm, as it matches the Hesai Pandar LiDAR \cite{HesaiXT32} in use, the mean octree population at $5cm$ resolution was found to be $~0.95$ points per voxel, consistent throughout the missions. This is also the same resolution used in state-of-the-art works, as shown in Tab. \ref{tab:octomap_resolution}. The resolution of an octree has a direct impact on its memory and CPU consumption; this choice of resolution was based on a careful evaluation of computational trade-offs, as evidenced by Fig. \ref{fig:octree_computational_cost}.

        % \begin{figure}[htbp]
        %     \centering
        %     \includegraphics[width=0.5\textwidth]{figures/octree_computational_cost.png}
        %     \caption[octree computational resource at different resolutions]{Evaluation of the computational resource at different octree resolutions of a the simulated LiDAR scan point cloud with 8,254,455 points. Lower memory usage, lower CPU time, and finer resolution is better. The octree was reconstructed at five different resolutions, and memory usage and mean CPU time over 10 iterations were recorded. Note; this was performed on a system with Intel Core i7-10875H CPU @ 2.30GHz $\times$ 16 cores, 32GB RAM.}
        %     \label{fig:octree_computational_cost}
        % \end{figure}

    \subsection{Segmentation and Filtering}
        The octree was then projected back onto the mission's point cloud map to give a \emph{difference point cloud} at the resolution of the original points. This process includes extracting the region of interest with a box filter and implementing a ground filter using RANSAC to remove ground points. We also smoothed the point cloud surfaces using Moving Least Squares (MLS) smoothing \cite{Lancaster1981SurfacesMethods}.
        Sensor noise in the object point clouds is filtered through erosion dilation to give a more accurate representation of the object point cloud.
        We then used Euclidean clustering to group points into distinct clusters, each potentially representing an object. To refine the clusters, we used region growing with normal estimation, allowing us to handle overlapping objects efficiently, as in \cite{FehrTSDF-basedDiscovery}.
        To better manage object overlap, we developed a custom method that can either merge or separate clusters based on their overlap ratio. This helps alleviate a common issue in direct model comparison, specifically, overlapping changes.

        These discrete object point clouds were then passed to an instance descriptor neural network.

    \subsection{Instance Descriptor Generation}
    \label{sec:inst_descriptor}
        %\mfallon{move the Instaloc section of the related work here. But leave a short reference to it in the related work saying we will describe the method in more detail in Section XX-X.}

        For feature description, we used a neural network for object instance description which generates a $16 \times 1$ SE(3) invariant descriptor for correspondence queries based on the method described in our previous work, InstaLoc~\cite{ZhangInstaLoc:Learning}.
        The descriptor network can generate object-level descriptors and can handle variations due to different viewpoints and partial observations.
        It employed a deep neural network to infer directly on 3D point cloud data using sparse tensors and spatially sparse convolutions for efficiency.
        In comparison, we saw that RIFT and SHOT descriptors did not scale well to large databases of objects due to their computational complexity, making the task of exhaustive comparison and matching objects in such databases resource intensive and time consuming.

        We adapt the InstaLoc descriptor network to determine object-level correspondences in multi-temporal data.
        We train the descriptor neural network on a large set of labeled point clouds, enabling it to generalize and perform well on unseen data. This is particularly useful for long-term autonomy in cluttered environments, as the robot can encounter unlabelled objects and still recognize them as changed, without the need for retraining. This is a significant advantage over traditional supervised methods that experience a domain gap and applications in one location does not transfer well to another.

    \subsection{Instance Grouping}
        % dim reduction clustering visualisation
        To determine the object correspondences between missions, the $16 \times 1$ descriptors are clustered.
        This was achieved by performing K-means clustering while automatically determining the most suitable number of clusters $K$ based on the elbow method, using Within Cluster Sum of Squares (WCSS) \cite{Kodinariya2013ReviewClustering}.
        % This method calculates the Within Cluster Sum of Squares (WCSS) to measure squared distances between data points and cluster centers.
        % The optimal $K$ is found at the 'elbow' of the WCSS curve, indicating no model improvement when another cluster is added.
        %The matches between objects in the simulated LiDAR are shown in the match matrix Fig. \ref{fig:object_match_matrix}.

        % If (optionally) using euclidean distancing clustering, looser clustering can be performed by increasing the distance threshold \cite{SoniMadhulatha2012ANMETHODS}, and more correspondences will be found between objects between missions.
        Each cluster has a confidence level based on the density of the cluster.
        Let $\mathbf{D}_1, \mathbf{D}_2, \ldots, \mathbf{D}_m$ be $m$ sets of feature descriptors, each representing a cluster, where $\mathbf{D}_i = \{d_{i1}, d_{i2}, \ldots, d_{in_i}\}$ contains $n_i$ feature descriptors, and each $d_{ij} \in \mathbb{R}^m$ is an $m$-dimensional feature descriptor.

        For each cluster $\mathbf{D}_i$, the centroid is given by:
        \begin{equation} \label{eq:centroid}
            \mathbf{C}_i = \frac{1}{n_i} \sum_{j=1}^{n_i} d_{ij}
        \end{equation}

        The average distance $\Bar{\Delta}$ of each descriptor in the cluster $\mathbf{D}_i$ to its centroid is calculated as:
        \begin{equation} \label{eq:distance}
            \Bar{\Delta}_i = \frac{1}{D} \sum_{d=1}^{D} \| d_{id} - \mathbf{C}_i \|_2
        \end{equation}

        Where $\| \cdot \|_2$ represents the $L_2$-norm (Euclidean norm), and $M$ is the number of dimensions of the data point $d_i$ and centroid $\mathbf{C}_i$, in our case $M=16$.

        The confidence metric for each cluster is then given by the min-max normalized distance, given by:
        \begin{equation}
               \text{cluster confidence}_i = \frac{\Bar{\Delta}_i - \Delta_{min}}{\Delta_{max} - \Delta_{min}}
        \end{equation}

        Where $\Delta_{min}$ is the minimum average distance among all clusters, and $\Delta_{max}$ is the maximum average distance among all clusters.

        Thus, for each cluster $\mathbf{D}_i$, cluster $\text{confidence}_i$ lies in the range $[0, 1]$, with $0$ indicating maximum compactness (all points close to the centroid) and $1$ indicating minimum compactness (high dispersion of points from the centroid).

        \subsubsection{Correspondence Confidence}
            % The confidence of each cluster was found using the silhouette plot method, as shown in Fig. \ref{fig:}, this can be used to further optimize $K$, as the average silhouette score is maximized.
            The next step is to determine the correspondences between objects intermission.
            If there are more than two instances of an object in the same class, exact intermission correspondences can only be estimated, so, we utilize a synthetic confidence metric for each object, a quantifiable measure to determine the normalized Euclidean distance between a data point $x_i$ and the centroid $c_j$ of its assigned cluster $C_j$.
            This ensures that data points closer to the centroid receive higher confidence scores, indicating a stronger alignment with the class. Then, classifications can be reported for the objects with confidence scores, as shown in the match matrix. Our synthetic confidence metric normalizes the confidence metric for each cluster based on the range of average distances across all clusters. Normalization allows for a consistent and comparable confidence metric across different clusters, even if they have different scales of average distances.
            The correspondences are then determined by minimizing the weighted normalized physical and descriptor distances, $\delta^{weighted}_{ij}$, as shown in Eq. \ref{eq:correspondence_weighting}, where $\delta^p$ is the distance in physical space, $\delta^d$ is the distance in descriptor space. The weights are $\alpha$ and $\beta$, respectively, and are user specifiable, such that if the space in which change is detected is small, and it is known that objects will not move far, then the physical distance can be up-weighted, and vice versa.

            \begin{equation}\label{eq:correspondence_weighting}
                \delta^{weighted}_{ij} = \alpha  \lVert \delta^p_{ij} \rVert_{\text{min-max-norm}} + \beta \lVert\delta^d_{ij} \rVert_{\text{min-max-norm}}
            \end{equation}

            An odd number of objects in a class implies additive or subtractive change, and so the object instance with the greatest weighted distance to other objects does not have a correspondence.
            % The results of which can be seen in \ref{synthetic confidence graph}.  % taken out for space...
        \subsubsection{Changed Object Registration}
            % find transformations between matched objects; assume same centroid, use svd to recover orientation
            For each point cloud representing a changed model instance in the scene, the correspondence grouping identifies the 6DOF pose estimation between the prior model and the instance of the model in the current scene.
            To determine the transformation between the corresponding objects,
            rigid point cloud registration technique was used. The translation
            is found by assuming the matched objects have the same centroid,
            and singular value decomposition (SVD) on the covariance matrix is
            used to recover the orientation \cite{Marden2012ImprovingSearch}.

\section{Experimental Results}

In this section, we present results and analysis from both simulated (Fig.
\ref{fig:3_topdown}) and real LiDAR mapping experiments (Fig.
\ref{fig:fsc_teaser}). With the
simulated LiDAR, every point is paired with its ground truth object label.
We use this later to measure change detection performance. The real-world
experiment was conducted in Fire Service College (FSC), an outdoor training
ground for firefighters. Our results demonstrate the system  working
robustly in both indoor and outdoor environments.

\subsection{Experiment with Simulated LiDAR Scans}

\subsubsection{Overview}
\begin{figure*}
	\centering
	\begin{subfigure}{0.32\textwidth}
		\centering
		\includegraphics[width=\textwidth]{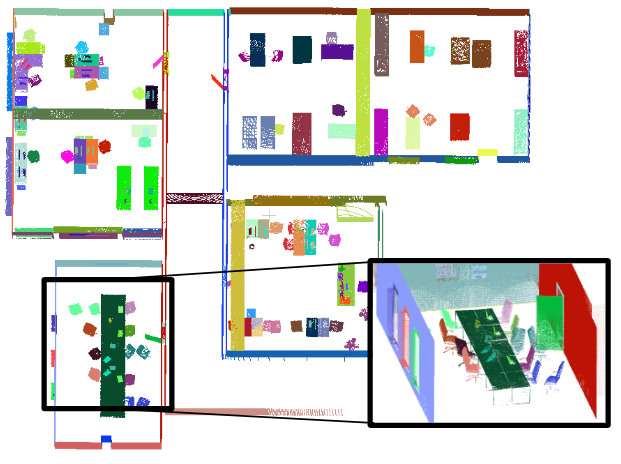}
		\caption{Simulated LiDAR dataset}
		\label{fig:static_sim}
	\end{subfigure}
	\hfill
	\begin{subfigure}{0.32\textwidth}
		\centering
		\includegraphics[width=\textwidth]{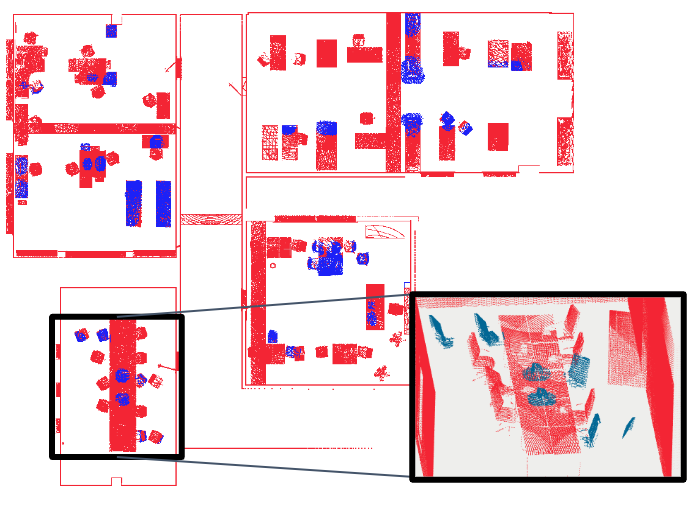}
		\caption{Identified intermission change}
		\label{fig:dynamic_sim}
	\end{subfigure}
	\hfill
	\begin{subfigure}{0.34\textwidth}
		\centering
		\includegraphics[width=\textwidth]{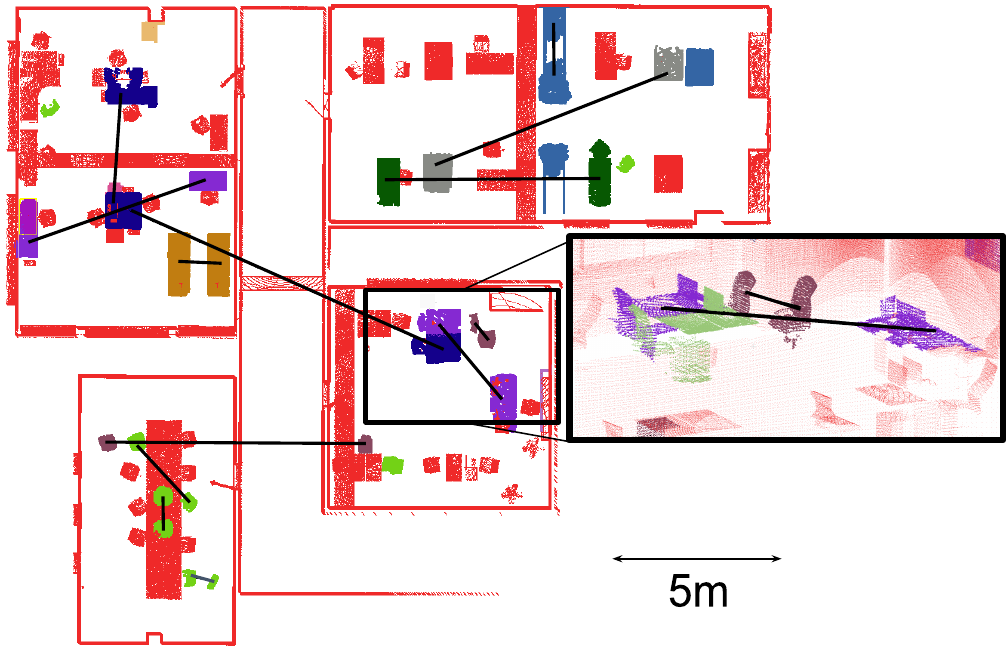}
		\caption{Object correspondences}
		\label{fig:correspondences_sim}
	\end{subfigure}
	\caption{Top down view of the change detection results of the simulated
		\emph{Exp. 1}. The ground and ceiling have been filtered out for
		visualization purposes. Fig. \ref{fig:dynamic_sim} shows segmented
		changed
		objects in blue, and the static prior map in red. Fig.
		\ref{fig:correspondences_sim} shows the most confident correspondences
		of objects which moved between missions.}
	\label{fig:3_topdown}
\end{figure*}

Using Unreal Engine 4 and AirSim \cite{Shah2018AirSim:Vehicles},
we generated three simulated LiDAR datasets in an office scene to
quantitatively evaluate our change detection methods. This allowed us to
determine metrics using the simulator ground truth (GT). The top-down
orthographic projection of one of the office scenes is shown in
Fig. \ref{fig:static_sim} , the
ground and ceiling are filtered out for visualization purposes. To
demonstrate our method can work on different LiDAR configurations, we
generated scans with the characteristics of the Ouster OS0-128 LiDAR which
has a field of view of 90 degrees and moved around following a route
similar to an inspection mission. Each dataset consists of two missions,
and the change detected between them was evaluated at a per-point level.

\begin{table}[h]
\centering
	\caption{\centering LiDAR change evaluation at a per-point level for the
	simulated experiments}
	\resizebox{\columnwidth}{!}{
	\begin{tabular}{c|c|c|c|c|c}
		\toprule
            \textbf{Exp.} & \textbf{Precision} & \textbf{Recall} &
            \textbf{Specificity} & \textbf{F-Score} & \textbf{IoU}\\
            \midrule
            $\mathbf{1}$ & $0.89$ & $0.68$  & $0.99$ & $0.77$ & $0.63$\\
            $\mathbf{2}$ & $0.82$ & $0.95$  & $0.99$ & $0.88$ &  $0.79$ \\
            $\mathbf{3}$ & $0.76$ & $0.55$ & $0.99$ & $0.63$ & $0.55$ \\
		\bottomrule
	\end{tabular}
	}
    \label{tab:sim_metrics}
\end{table}

\subsubsection{Change Detection}
We evaluated precision, recall, and specificity to understand the performance
of the change detection algorithm. The results are shown in Tab.
\ref{tab:sim_metrics}. On average across three datasets, we achieve a precision
of $0.82$ and a recall of $0.73$.
%  The overlap between GT and detected change (as shown in Fig.
%\ref{fig:dynamic_sim}) was determined using a KD-Tree
%\cite{Hapala2011Review:Tracing}.  \mfallon{you have said per-point twice here}
Note that these metrics are partly dependent on the consistency of the raw
observations intermission. \emph{Exp. 3} exhibited lower recall value due to
many overlapping object changes intermission, causing incomplete segmentation. 
Whereas \emph{Exp. 2} had fewer overlapping changed objects.

% If a changed object is partially occluded in one of the missions, then the
%segmentation will only be partial. The recall can be improved by having more
%consistent observations of changed objects throughout the missions. However,
%the per-point identified changes are complete enough for robust description by
%the learned point cloud descriptor network.

% \emph{Exp. 3} had fewer changed objects than \emph{Exp. 1 \& 2}, and so the
%recall was more heavily influenced by the individual changes, making it
%particularly sensitive to the accuracy of the per-point change detection.
% \mfallon{comment: but the value of the method is if it can detect change
%without obsessively scanning and rescanning a room. (why is recall lower in
%exp3) Fixed-JR }

\subsection{Experiment with Real World LiDAR Scans}

\subsubsection{Overview}

As shown in Fig. \ref{fig:fsc_teaser}, our system is intended to be used by a
Boston Dynamics Spot robot performing routine inspections in an industrial
facility. The sensor payload comprises a 3D LiDAR, multiple cameras, and a
self-contained IMU, as shown in Fig. \ref{fig:fsc_teaser}. In particular, we
use the Hesai PandarXT-32 LiDAR. %$$\footnote{XT32, Mid-Range Mechanical Lidar,
%HESAI Technology,
%url = {\url{https://www.hesaitech.com/product/xt32/}}}.
Real world data was collected using this platform at the Fire Service College
(FSC), where there were a variety of atypical objects, e.g. crates of furniture
and wood.
Multiple missions were executed where the robot autonomously followed a
predefined path. Typically, the trajectory executed by the robot fell within
\SI{20}{\centi\meter} of the desired trajectory, making the experiments highly
repeatable.

\subsubsection{Change Detection}
\label{sec:fsc_change_detection}
\begin{figure}
	\centering
	\includegraphics[width = 0.9\columnwidth]{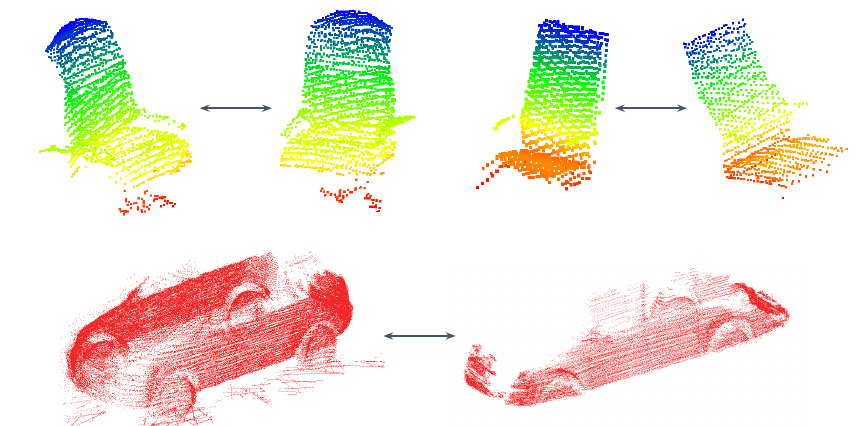}
	\caption{Examples of segmented object point cloud clusters of the same
		class, after unsupervised classification. Top: chairs from the
		simulated
		LiDAR dataset. Bottom: cars from the real-world FSC experiment. This
		shows
		the robustness of the classifier to minor occlusion and different
		viewpoints.}
	\label{fig:segmented_objects}
\end{figure}

Two of the missions registered using multi-mission pose-graph optimization are
shown in Fig. \ref{fig:fsc_teaser}, showing that the multi-mission SLAM makes
the system robust to subtle inaccuracies within the point cloud maps. The
changed objects identified from the prior mission are highlighted in red. A
further example of a changed object is shown at the bottom of Fig.
\ref{fig:segmented_objects}. Our approach retains the capability to establish 
correspondences for partially observed objects, demonstrating the robustness of both the descriptor and the classifier.

\begin{figure}
    \centering
    \includegraphics[width=\columnwidth]{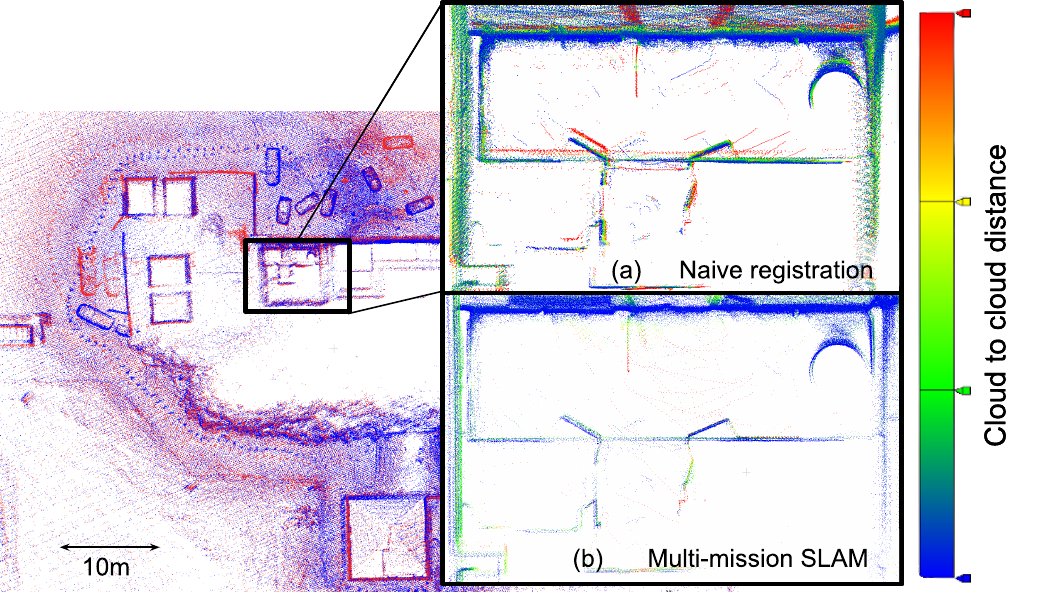}
    \caption{Comparison between (a) naive direct ICP alignment and (b) our
    multi-mission SLAM method. (a) Shows the cloud-to-cloud distance of a local
    part of the mild misregisteration between two rigid global point clouds,
    ``double walling'' occurs locally causing phantom change to be detected.
    (b) Shows the cloud to cloud distance of our multi-mission registration
    solving this problem.}
    \label{fig:global_misreg}
 \end{figure}

Alternatively, if the missions had been registered using ICP on the full global
maps from each mission, some of the points would not be well aligned due to
imprecision in the SLAM. As a direct comparison, we used ICP to register two
global maps from two FSC missions, shown in Fig. \ref{fig:global_misreg}. As
shown in Fig. \ref{fig:global_misreg} (a) and (b), this misregistration can
then cause erroneous change detection. Furthermore, the sensor origin is also
needed to produce correct representation of free space using octrees.

\subsubsection{Instance Grouping}
\begin{figure}
	%\centering
	\includegraphics[width=0.9\columnwidth]{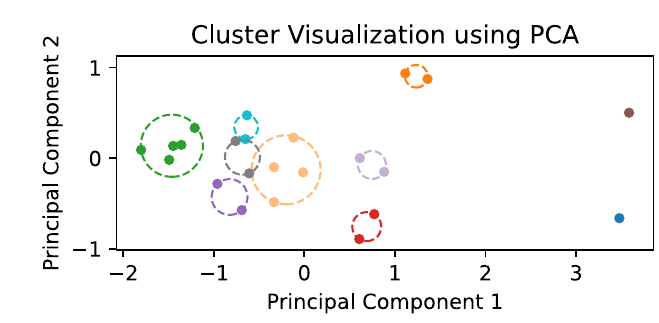}
	\caption{Object feature descriptors from FSC experiment, reduced from $16
		\times 1$ to $2$ dimensions for visualization using PCA.}
	\label{fig:cluster_vis}
\end{figure}

The classified object feature descriptors are shown in Fig.
\ref{fig:cluster_vis}, with Principal Component Analysis (PCA) used for
dimensionality reduction (from 16) for ease of presentation. This is presented
to show that our clustering algorithm can effectively group the high
dimensional point cloud descriptors.

Fig. \ref{fig:segmented_objects} shows instances of one of the changed objects,
after classification of learned descriptors.

\begin{figure}
	\centering
	\includegraphics[width =
	0.7\columnwidth]{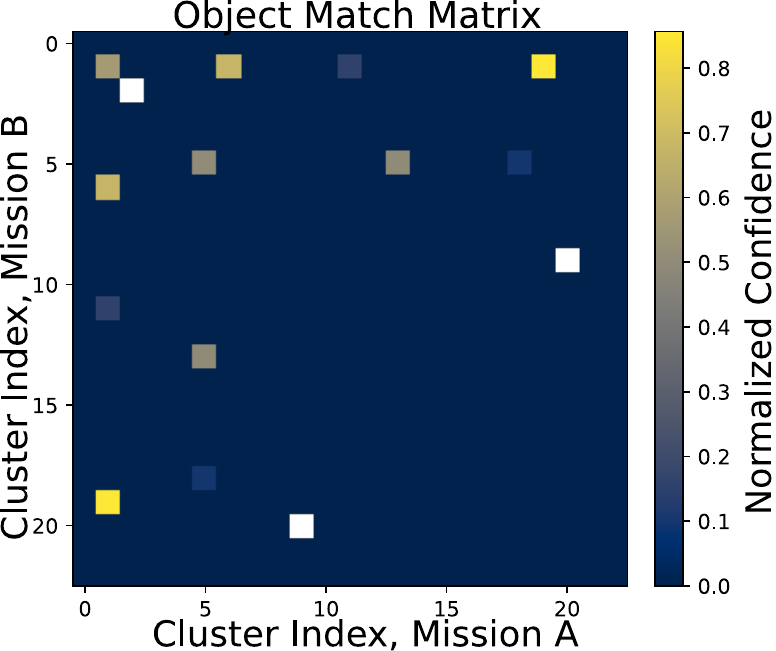}
	\caption{Object match matrix for FSC experiment, showing correspondences
		between objects intermission, with heatmap for the confidence of
		classification, based on the normalized distance of cluster to
		centroid.}
	\label{fig:spot_fsc_match_matrix}
\end{figure}

In particular, this figure illustrates that incomplete observations do not
affect the performance of the classifier.
The match matrix for classified objects is shown in Fig.
\ref{fig:spot_fsc_match_matrix}, where the colormap corresponds to the
confidence of the classification, based on the descriptor cluster density. In
this figure, all potential matches are shown for each object within the class.

As a comparison, we ran SHOT and RIFT descriptors on $150$ objects after
down-sampling point clouds to \SI{5}{\centi\meter}. Their matching performance
is poor as SHOT and RIFT are per-point descriptors, which means that we need to
find per-point correspondences for each object to determine if two objects are
similar. As importantly, it took \SI{4.7}{\second} for SHOT and
\SI{1.2}{\second} for RIFT for each object. In contrast, our descriptor network
is much faster, taking only about \SI{10}{\milli\second} for each object.

\subsection{Implementation Details}
These experiments were performed on a system with Intel Core i7-10875H CPU @
2.30GHz $\times$ 16 cores, 32GB RAM. For the simulated LiDAR office scene
\emph{Exp. 1}, the octree reconstruction and differencing took
$\sim$\SI{60}{\second}, segmentation and filtering $\sim$\SI{7}{\second}, and
instance description, grouping and objects registration $<$\SI{1}{\second}.

It took between 10 and \SI{12}{\minute} for the Spot robot to execute each FSC
mission, which was \SI{250}{\meter} in length.
% \mfallon{which was XXm in length}.
For the FSC experiments octree reconstruction and differencing took
$\sim$\SI{6}{\minute}, segmentation and filtering $\sim$\SI{18}{\second}, and
instance description, grouping and object registration $<$\SI{2}{\second}.

\subsection{Challenges/Assumptions}
\begin{enumerate}
    \item Objects are assumed to be rigid, non-deformable, and spatially
    coherent – assuming that nearby points belong to the same distinct object.

    \item If multiple instances of the same object class are moved between
    missions, identical objects could be confused. The estimated
    correspondences are determined by maximising the confidence in the
    descriptor clustering.

    \item We assume that the environment is sufficiently consistent that the
    SLAM missions can be registered according to the method described in Sec.
    \ref{sec:multi-mission}

    \item Objects which move by less than the size of the object present a
    challenge for change detection algorithms as part of the physical space is
    still overlapping. This causes pairwise cloud differencing to fail and to
    merge clouds. This merging forms larger composite segments, concealing
    complete changes at the object level.

\end{enumerate}

\section{Conclusion}
In this paper, we proposed LiSTA, an object-level LiDAR change detection
approach. LiSTA can be used to determine the semi-static changes between
successive SLAM missions, and report the correspondences between the changed
objects intermission. LiSTA can determine correspondences without requiring
pre-training on a dataset of known objects. Our method provides effective
segmentation and object extraction capabilities from point cloud data. In
future, we aim to use color/visual information to further improve segmentation
and correspondence grouping. Lastly, we aim to implement octree submapping to
allow for scalability to very large scenes and real-time operation - enabling
our quadruped's autonomy system to re-plan its missions online when change is
identified.

\section{Acknowledgment}

We acknowledge Ren Komatsu (University of Tokyo) for his contribution in the early stages of this
project and Tobit Flatscher for his help with the real-world experiments.

\bibliographystyle{ieeetr}
\bibliography{references_new.bib}
\end{document}